\documentclass[11pt]{article}

\usepackage[margin=1in]{geometry}
\usepackage[T1]{fontenc}
\usepackage{amsmath,amssymb,mathtools}
\usepackage{booktabs}
\usepackage{multirow}
\usepackage{graphicx}
\usepackage{subcaption}
\usepackage{algorithm}
\usepackage{algpseudocode}
\usepackage{xspace}
\usepackage[dvipsnames]{xcolor}
\usepackage{enumitem}
\usepackage{microtype}
\usepackage{siunitx}
\usepackage{pifont}
\usepackage{url}
\usepackage{wrapfig}
\usepackage{float}
\usepackage[square,numbers,sort&compress]{natbib}
\usepackage[hidelinks]{hyperref}

\title{Permutation Robustness Is Not Enough: Action Collapse in Multi-Agent Transformer Policies}
\author{%
Amit Thakur \qquad Mukesh Singhal\\[0.35em]
\small University of California, Merced
}
\date{}

\hypersetup{
  pdftitle={Permutation Robustness Is Not Enough: Action Collapse in Multi-Agent Transformer Policies},
  pdfauthor={Amit Thakur and Mukesh Singhal},
  pdfkeywords={multi-agent reinforcement learning, transformer policies, permutation equivariance, action collapse, robot learning}
}

\newcommand{\E}{\mathbb{E}}
\newcommand{\R}{\mathbb{R}}

\newcommand{\modelid}{\textsc{PPO-ID}\xspace}
\newcommand{\modeleq}{\textsc{PPO-ID+Eq}\xspace}

\begin{document}
\maketitle

\begin{abstract}
Transformer policies are attractive for multi-agent robot learning because self-attention can model interactions among agents. However, multi-agent teams are unordered, while transformers typically process agents as ordered token sequences. We study how this mismatch affects cooperative navigation policies under agent-order permutations. Our results show that low permutation error alone can be misleading: policies may appear robust simply because all agents choose the same action. We therefore evaluate policies using both permutation-consistency metrics and action-collapse diagnostics, including action diversity, same-action fraction, and maximum action frequency. A PPO-ID baseline yields non-collapsed behavior but remains order-sensitive, while strong equivariance regularization can still induce homogeneous behavior. A weak equivariance penalty improves the robustness while preserving more diverse actions for teams with \(N=3\) agents, whereas teams with \(N=4\) agents require substantially smaller regularization weights. These findings suggest that multi-agent transformer policies should be evaluated not only by return and permutation robustness, but also by whether they maintain non-collapsed, differentiated multi-agent behavior.
\end{abstract}

\begin{center}
\end{center}

\section{Introduction}
Multi-robot systems are naturally composed of exchangeable entities. A team of ground robots, drones, or manipulation agents should not fail because the observations of two robots are listed in a different order. Yet transformer policies, now common as interaction models, operate on ordered token sequences~\citep{vaswani2017attention}. A typical multi-agent transformer forms tokens
\begin{equation}
    X=[x_1,\ldots,x_N],\qquad x_i=f_{\rm emb}(o_i),
\end{equation}
where $o_i$ is agent $i$'s observation. If the same physical team is represented by $PX$, where $P$ is a permutation matrix, the desired behavior is the equivariance:
\begin{equation}
    f(PX)=P f(X). \label{eq:intro_equiv}
\end{equation}
This condition says that shuffling the input agents should only shuffle the corresponding output commands.

Practical policies often add learned agent identity or sequence-position embeddings to the tokens. Such embeddings can improve nominal performance because agents may specialize, but they can also create an ordering shortcut: the policy attaches behavior to token slots rather than to physical context. This is problematic because robot ordering may change due to communication, batching, logging, dynamic team membership, or a deployment interface.

In this paper, we address the following question: \emph{do transformer-based multi-agent policies learn coordination or agent ordering?} A natural test is to measure Eq.~\eqref{eq:intro_equiv}. However, our experiments reveal that low equivariance error can be misleading. A policy can satisfy permutation tests almost perfectly by outputting the same action for every agent. Such a policy can appear permutation-consistent for the wrong reason: not because it coordinates robustly, but because all agents behave nearly identically. We therefore evaluate task return, permutation consistency, and non-collapsed action behavior jointly.

Our contributions are as follows. (i) We formalize permutation sensitivity for transformer-based multi-agent policies with logit, policy-distribution, action, and return metrics. (ii) We introduce action-collapse diagnostics, including normalized action diversity, same-action fraction, and maximum action frequency. (iii) We show empirically that PPO-style training~\citep{schulman2017ppo,yu2022mappo} produces a strong non-collapsed baseline, but equivariance regularization still creates a measurable robustness-diversity tradeoff. (iv) We add a four-agent stress test showing that useful regularization strength is team-size dependent.

The practical message is deliberately conservative: a low permutation error should not be interpreted in isolation. A good multi-agent policy should satisfy three requirements simultaneously. It should solve the task, remain stable under benign relabelings of the same physical team, and avoid degenerate homogeneous behavior. The experiments below are organized around these three requirements.

\section{Related Work}
\paragraph{Cooperative multi-agent reinforcement learning:} 
Cooperative MARL studies how a team of agents can optimize a shared objective under coupled dynamics and partial observations~\citep{stone2000multiagent,sutton2018reinforcement}. Centralized-training methods such as MADDPG~\citep{lowe2017maddpg}, COMA~\citep{foerster2018coma}, VDN~\citep{sunehag2018vdn}, and QMIX~\citep{rashid2018qmix} address credit assignment and coordination by using centralized information during learning. PPO-based multi-agent training has also emerged as a strong practical baseline in cooperative games~\citep{yu2022mappo}. Our work uses this PPO/MAPPO-style training regime, but focuses on a different question: whether a transformer policy's representation of agents is stable to arbitrary ordering.

\paragraph{Transformers, sets, and symmetry:}
Transformers~\citep{vaswani2017attention,devlin2019bert} process inputs as token sequences. Without position-dependent information, self-attention is permutation-equivariant over tokens; learned position or agent-ID embeddings generally break that property. Set architectures such as Deep Sets~\citep{zaheer2017deepsets} and Set Transformer~\citep{lee2019settransformer} explicitly enforce permutation symmetry, and geometric deep learning emphasizes matching neural architectures to the symmetries of the domain~\citep{bronstein2021geometric}. Robot teams have a natural exchangeability symmetry: changing the list order of identical agents should not change the physical command assigned to each robot.

\paragraph{Attention for coordination:}
Attention has been used for communication, critic design, and interaction modeling in MARL~\citep{jiang2018attncomm,iqbal2019actorattention}. Emergent cooperative behavior and role differentiation have also been studied in multi-agent populations~\citep{mordatch2017emergence,baker2020emergent,wang2020maven}. These works motivate attention as a useful inductive bias for interaction, but most evaluations emphasize return. Our results show that return can hide order sensitivity, while equivariance error can hide same-action collapse. The proposed diagnostics are meant to complement rather than replace standard task performance.

\paragraph{Evaluation beyond return:}
The core methodological point of this paper is that robustness metrics can be gamed by degenerate policies. In our setting, a policy that emits the same logits for every agent is perfectly permutation-consistent, yet it may not represent meaningful coordination. This failure mode is analogous to shortcut solutions in representation learning: the metric is optimized, but the intended behavior is not obtained. We therefore evaluate distributional consistency, executed-action consistency, and action diversity together.

\section{Problem Formulation}
We consider a cooperative partially observable Markov game
\begin{equation}
\mathcal{G}=\big(\mathcal{S},\{\mathcal{A}_i\}_{i=1}^{N},P_{\rm env},r,\{\mathcal{O}_i\}_{i=1}^{N},\gamma\big).
\end{equation}
At time $t$, the environment state is $s_t\in\mathcal{S}$, agent $i$ receives observation $o_i^t\in\mathcal{O}_i$, the joint observation is $O_t=[o_1^t,\ldots,o_N^t]$, and the joint action is $a_t=(a_1^t,\ldots,a_N^t)$. The shared reward is $r_t=r(s_t,a_t)$, and the objective is
\begin{equation}
    J(\theta)=\E_{\pi_\theta}\left[\sum_{t=0}^{T}\gamma^t r_t\right].
\end{equation}

\paragraph{Notation:}
$N$ is the number of agents, $i\in\{1,\ldots,N\}$ indexes an agent, and $t\in\{0,\ldots,T\}$ indexes time within an episode of horizon $T$. The state space is $\mathcal{S}$, $\mathcal{A}_i$ is the action space of agent $i$, $\mathcal{O}_i$ is its observation space, and $P_{\rm env}$ denotes the environment transition kernel. The scalar $\gamma\in[0,1]$ is the discount factor, which controls the relative weight of future rewards. The symbol $\theta$ denotes all learned neural-network parameters, and $\pi_\theta$ is the stochastic policy induced by those parameters. The expectation $\E_{\pi_\theta}$ is taken over trajectories generated by executing $\pi_\theta$ in the environment.

\paragraph{Transformer policy:}
Each observation is mapped to a $d$-dimensional token $x_i=f_{\rm emb}(o_i)\in\R^d$, where $d$ is the transformer hidden dimension. With learned identity and positional embeddings, the input token is
\begin{equation}
    h_i=x_i+e_i+p_i,
\end{equation}
where $e_i\in\R^d$ is the learned embedding associated with agent identity and $p_i\in\R^d$ is the learned embedding associated with sequence position. A transformer encoder maps the input token sequence to contextualized output tokens
\begin{equation}
    H=[H_1,\ldots,H_N]=\mathrm{Transformer}(h_1,\ldots,h_N),
\end{equation}
where $H_i\in\R^d$ is the output representation used to choose the action of agent $i$. The action head is a linear classifier with weights $W$ and bias $b$, producing logits
\begin{equation}
    z_i=WH_i+b\in\R^{|\mathcal{A}|},
\end{equation}
where $|\mathcal{A}|$ is the number of discrete actions. The stochastic policy for agent $i$ is
\begin{equation}
    \pi_\theta(a_i\mid O_t)=\mathrm{softmax}(z_i)_{a_i}.
\end{equation}
We use a factorized joint policy conditioned on all observations,
\begin{equation}
    \pi_\theta(a_t\mid O_t)=\prod_{i=1}^{N}\pi_\theta(a_i^t\mid O_t),\qquad
    \log \pi_\theta(a_t\mid O_t)=\sum_i\log\pi_\theta(a_i^t\mid O_t).
\end{equation}
Although the joint policy factorizes over sampled actions, each factor can still depend on all agents through the transformer attention computation over $O_t$.

\paragraph{Permutation sensitivity:}
Let $X=[h_1,\ldots,h_N]$ denote the ordered matrix of input tokens, and let $f_\theta(X)$ denote the stacked action logits $Z\in\R^{N\times |\mathcal{A}|}$. A permutation matrix $P\in\{0,1\}^{N\times N}$ reorders the agent dimension. A policy is permutation-equivariant if
\begin{equation}
    f_\theta(PX)=P f_\theta(X). \label{eq:equivariance}
\end{equation}
It means that if the input agent order is permuted, the output logits should be permuted in exactly the same way. If identity or position embeddings are tied to sequence slots, then moving a physical agent to a new slot changes its embedding and Eq.~\eqref{eq:equivariance} may not hold.

\section{Diagnostics and Training Objective}
\paragraph{Permutation metrics:}
We report mean return under normal, random, and reverse ordering. We compare normal logits to permuted logits after unpermuting the outputs:
\begin{equation}
    E_{\rm eq}(X,P)=\left\|P^{-1}f_\theta(PX)-f_\theta(X)\right\|_2^2. \label{eq:eqerr}
\end{equation}
We also compute policy KL divergence
\begin{equation}
    D_{\rm KL}\left(\pi(\cdot\mid X)\,\|\,P^{-1}\pi(\cdot\mid PX)\right)
\end{equation}
and greedy action disagreement
\begin{equation}
    D_{\rm act}=\frac{1}{N}\sum_i \mathbf{1}\left[\arg\max_a\pi_i(a\mid X)\neq \arg\max_a (P^{-1}\pi(PX))_i(a)\right].
\end{equation}

\paragraph{Collapse metrics:}
For selected greedy actions $a_1^t,\ldots,a_N^t$, normalized action diversity is
\begin{equation}
    D_{\rm div}=\frac{1}{T}\sum_{t=1}^{T}\frac{|\{a_1^t,\ldots,a_N^t\}|}{N}. \label{eq:diversity}
\end{equation}
Complete same-action collapse has $D_{\rm div}=1/N$. We also report same-action fraction
\begin{equation}
    F_{\rm same}=\frac{1}{T}\sum_t \mathbf{1}[a_1^t=\cdots=a_N^t]
\end{equation}
and maximum empirical action frequency $F_{\max}=\max_a c_a/\sum_{a'}c_{a'}$, where $c_a$ counts action $a$ across agents and timesteps.

\paragraph{PPO/MAPPO-style training:}
We use a centralized transformer actor and centralized critic. PPO uses the probability ratio
\begin{equation}
    \rho_t(\theta)=\exp\left(\log\pi_\theta(a_t\mid O_t)-\log\pi_{\theta_{\rm old}}(a_t\mid O_t)\right),
\end{equation}
and clipped objective
\begin{equation}
    L_{\rm clip}=-\E_t\left[\min\left(\rho_t\hat{A}_t,\mathrm{clip}(\rho_t,1-\epsilon,1+\epsilon)\hat{A}_t\right)\right],
\end{equation}
with GAE advantages~\citep{schulman2016gae}. We add equivariance regularization
\begin{equation}
    L_{\rm eq}=\left\|P^{-1}f_\theta(PX)-f_\theta(X)\right\|_2^2,
\end{equation}
and optionally pairwise policy diversity
\begin{equation}
    L_{\rm pair}=-\frac{1}{N(N-1)}\sum_{i\neq j}\|\pi_i-\pi_j\|_2^2.
\end{equation}
We also use a standard entropy bonus to prevent the policy from becoming deterministic too early. The average policy entropy is
\begin{equation}
    H(\pi_\theta)
    =\E_t\left[
    \frac{1}{N}\sum_{i=1}^{N}
    \left(-\sum_{a\in\mathcal{A}}\pi_\theta(a\mid O_t)\log \pi_\theta(a\mid O_t)\right)
    \right].
    \label{eq:entropy}
\end{equation}
Equivalently, the entropy loss is \(L_{\rm ent}=-H(\pi_\theta)\); thus the term \(-c_HH(\pi_\theta)\) in the minimized objective encourages exploration and discourages premature deterministic action collapse. 

The value loss trains the centralized critic to predict the rollout return:
\begin{equation}
    L_V
    =
    \mathbb{E}_t
    \left[
    \left(
    V_\theta(O_t)-\hat{R}_t
    \right)^2
    \right],
    \label{eq:value-loss}
\end{equation}
where \(V_\theta(O_t)\) is the critic prediction for the joint observation and
\(\hat{R}_t\) is the empirical or GAE-based return target. A more accurate critic
reduces variance in the policy-gradient update by producing better advantage
estimates.

The complete objective is
\begin{equation}
    L=L_{\rm clip}+c_vL_V-c_HH(\pi_\theta)+\lambda_{\rm eq}L_{\rm eq}+\lambda_{\rm pair}L_{\rm pair}. \label{eq:loss}
\end{equation}

\begin{algorithm}[t]
\caption{PPO-style permutation-regularized multi-agent training}
\label{alg:ppo}
\begin{algorithmic}[1]
\State Initialize transformer actor-critic $f_\theta,V_\phi$
\For{each PPO update}
    \State Collect rollouts and store $O_t,a_t,r_t,\log\pi_{\theta_{old}}(a_t\mid O_t),V_\phi(O_t)$
    \State Compute GAE advantages $\hat A_t$ and returns
    \For{each minibatch $X$}
        \State Compute PPO clipped loss and value loss
        \State Sample a permutation $P$ and compute $L_{\rm eq}=\|P^{-1}f_\theta(PX)-f_\theta(X)\|_2^2$
        \State Optionally compute $L_{\rm pair}$ from pairwise action distributions
        \State Calculate entropy loss $L_{ent}$
        \State Update parameters with Eq.~\eqref{eq:loss}
    \EndFor
\EndFor
\State Evaluate normal, random, and reversed ordering with performance, robustness, and collapse metrics.
\end{algorithmic}
\end{algorithm}

\section{Experimental Setup}
We use the PettingZoo/MPE cooperative navigation task~\citep{terry2021pettingzoo,mordatch2017emergence}. In this task, $N$ agents move in a shared plane and receive a cooperative reward that favors covering $N$ landmarks while avoiding inefficient configurations. The task is simple enough to permit controlled perturbation tests, but it still requires agents to condition their behavior on the relative configuration of other agents and landmarks. This makes it a useful diagnostic environment for distinguishing three phenomena that are otherwise conflated by return alone: order sensitivity, useful specialization, and homogeneous action collapse.

\paragraph{Architectures and training:}
The transformer policy embeds each agent observation into a 128-dimensional token. The policy uses two self-attention blocks with four heads, learned agent ID embeddings, and learned sequence-position embeddings. The actor head maps each final token to logits over discrete actions. The critic is centralized and estimates a team value from the joint representation. We use PPO-style multi-agent training with rollout length 1024, four PPO epochs, minibatch size 256, learning rate $3\times10^{-4}$, discount $\gamma=0.99$, GAE parameter $0.95$, PPO clip $0.2$, entropy coefficient $0.01$, value coefficient $0.5$, and gradient clipping at $0.5$. The policy factorizes across agents, but each factor is conditioned on the full set of agent observations through the transformer.

\paragraph{Ordering interventions:}
For every trained model, evaluation is run in three modes. In the \emph{normal} mode, observations are provided in the canonical environment order. In the \emph{random} mode, a fresh deterministic random permutation is applied to the agent-token order before each evaluation episode. In the \emph{reverse} mode, the order is reversed. After a permuted forward pass, output logits and selected actions are unpermuted back to the original physical-agent order before computing metrics or stepping the environment. This procedure isolates ordering sensitivity from changes in the physical state.

\paragraph{Compared variants:}
For the main $N=3$ comparison, we report mean and standard deviation over three seeds. We compare: \modelid, a PPO transformer with ID/position embeddings; \modeleq with permutation augmentation and weak equivariance regularization $\lambda_{\rm eq}=0.01$; stronger regularization $\lambda_{\rm eq}=0.05$; and $\lambda_{\rm eq}=0.05$ with pairwise policy-diversity regularization $\lambda_{\rm pair}=0.007$. The weak coefficient is selected because stronger coefficients reduce equivariance error but also push the policy toward same-action behavior. We therefore select hyperparameters using a joint criterion: high return, low permutation error, and non-collapsed action statistics.

\paragraph{Team-size stress test:}
We also evaluate $N=4$ agents and four landmarks as a scalability stress test. This setting increases the number of possible orderings from $3!=6$ to $4!=24$ and lowers the collapse baseline from $D_{\rm div}=1/3$ to $D_{\rm div}=1/4$. Because preliminary $N=4$ runs show stronger collapse under the $N=3$ regularizer, we sweep much smaller coefficients, including $2.5\times10^{-4}$, $5\times10^{-4}$, $10^{-3}$, $10^{-2}$, and $5\times10^{-2}$. The $N=4$ results are reported as a seed-0 stress test rather than a full benchmark, and are used to assess whether the qualitative tradeoffs persist as team size grows.

\section{Results}
\begin{figure*}[t]
\centering
\includegraphics[width=\textwidth]{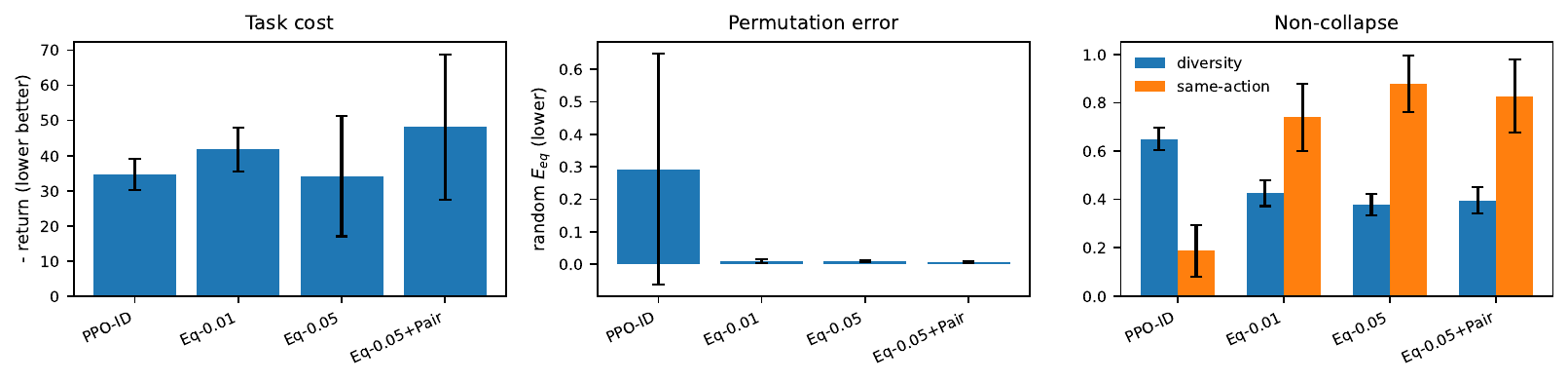}
\caption{Three-agent PPO results. The ID/position baseline is diverse but order-sensitive. Weak equivariance regularization improves permutation consistency but reduces diversity; stronger regularization moves toward homogeneous behavior.}
\label{fig:n3_tradeoff}
\end{figure*}

\begin{table*}[t]
\centering
\small
\caption{Main PPO results with $N=3$ agents, mean $\pm$ standard deviation across three seeds. Return is episode return; $E_{\rm eq}^{\rm rand}$, KL, and $D_{\rm act}$ are computed under random permutations. $D_{\rm div}$ and $F_{\rm same}$ are computed under normal ordering.}
\label{tab:n3}
\resizebox{\textwidth}{!}{%
\begin{tabular}{lcccccc}
\toprule
Method & Return & $E_{\rm eq}^{\rm rand}$ & KL$^{\rm rand}$ & $D_{\rm act}^{\rm rand}$ & $D_{\rm div}$ & $F_{\rm same}$ \\
\midrule
\modelid & $-34.67\pm4.45$ & $0.293\pm0.355$ & $0.117\pm0.147$ & $0.498\pm0.052$ & $0.649\pm0.047$ & $0.187\pm0.108$ \\
\modeleq, $\lambda_{\rm eq}=0.01$ & $-41.75\pm6.24$ & $\mathbf{0.009}\pm0.006$ & $\mathbf{0.002}\pm0.001$ & $0.196\pm0.103$ & $0.425\pm0.053$ & $0.739\pm0.139$ \\
Eq, $\lambda_{\rm eq}=0.05$ & $-34.13\pm17.13$ & $\mathbf{0.009}\pm0.003$ & $0.003\pm0.000$ & $\mathbf{0.112}\pm0.102$ & $0.378\pm0.043$ & $0.878\pm0.117$ \\
Eq+Pair, $\lambda_{\rm eq}=0.05,\lambda_{\rm pair}=0.007$ & $-48.16\pm20.70$ & $\mathbf{0.007}\pm0.004$ & $\mathbf{0.002}\pm0.001$ & $0.150\pm0.135$ & $0.396\pm0.054$ & $0.826\pm0.151$ \\
\bottomrule
\end{tabular}}
\end{table*}

\paragraph{PPO-ID yields non-collapsed behavior, but remains order-sensitive:}
Table~\ref{tab:n3} and Fig.~\ref{fig:n3_tradeoff} show that \modelid obtains the strongest non-collapsed behavior among the $N=3$ PPO variants: $D_{\rm div}=0.649\pm0.047$ and $F_{\rm same}=0.187\pm0.108$. However, it remains order-sensitive: random action disagreement is $0.498\pm0.052$. Thus, PPO-ID provides a useful non-collapsed baseline, but learned ID/position embeddings remain vulnerable to permutation changes.

\paragraph{Equivariance regularization improves robustness but reduces diversity:}
With $\lambda_{\rm eq}=0.01$, random equivariance error falls from $0.293$ to $0.009$, and action disagreement falls from $0.498$ to $0.196$. However, action diversity also falls from $0.649$ to $0.425$ and same-action fraction rises from $0.187$ to $0.739$. This is the central tradeoff: regularization improves permutation consistency but can encourage homogeneous behavior.

\paragraph{Low equivariance error alone is insufficient:}
The stronger $\lambda_{\rm eq}=0.05$ and Eq+Pair settings have excellent permutation metrics but high same-action fractions. These models would look strong if evaluated only with Eq.~\eqref{eq:eqerr}, but collapse diagnostics reveal a different story. Pairwise policy diversity helps only partially in this setting, which suggests that future methods should use task-aware diversity or relative geometric attention rather than blind action separation.

\begin{figure*}[t]
\centering
\includegraphics[width=\textwidth]{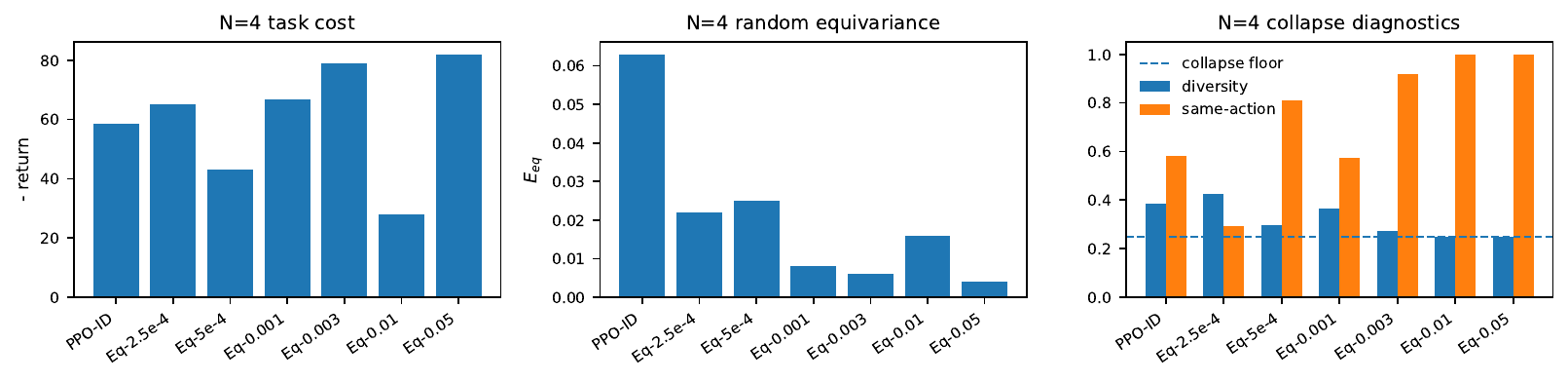}
\caption{Four-agent stress test. The collapse floor is $D_{\rm div}=1/4=0.25$. Regularization strengths that are useful for $N=3$ can fully collapse $N=4$ policies, while much weaker values reveal a return-diversity tradeoff.}
\label{fig:n4}
\end{figure*}

\begin{table*}[t]
\centering
\small
\caption{Four-agent stress test, seed 0. The coefficient that is useful for $N=3$ ($\lambda_{\rm eq}=0.01$) fully collapses for $N=4$. Smaller coefficients expose a tradeoff between return, robustness, and non-collapse.}
\label{tab:n4}
\resizebox{0.7\textwidth}{!}{%
\begin{tabular}{lcccccc}
\toprule
Method & Return & $E_{\rm eq}^{\rm rand}$ & $D_{\rm act}^{\rm rand}$ & $D_{\rm div}$ & $F_{\rm same}$ & $F_{\max}$ \\
\midrule
PPO-ID & $-58.43$ & $0.063$ & $0.331$ & $0.384$ & $0.582$ & $0.719$ \\
Eq, $\lambda_{\rm eq}=2.5\!\times\!10^{-4}$ & $-64.99$ & $0.022$ & $0.369$ & $\mathbf{0.427}$ & $\mathbf{0.293}$ & $\mathbf{0.591}$ \\
Eq, $\lambda_{\rm eq}=5\!\times\!10^{-4}$ & $\mathbf{-43.12}$ & $0.025$ & $0.121$ & $0.297$ & $0.811$ & $0.704$ \\
Eq, $\lambda_{\rm eq}=10^{-3}$ & $-66.65$ & $\mathbf{0.008}$ & $0.310$ & $0.367$ & $0.575$ & $0.751$ \\
Eq, $\lambda_{\rm eq}=10^{-2}$ & $-28.04$ & $0.016$ & $0.000$ & $0.250$ & $1.000$ & $1.000$ \\
Eq, $\lambda_{\rm eq}=5\!\times\!10^{-2}$ & $-82.00$ & $0.004$ & $0.000$ & $0.250$ & $1.000$ & $1.000$ \\
\bottomrule
\end{tabular}}
\end{table*}

\paragraph{Team-size sensitivity:}
Table~\ref{tab:n4} and Fig.~\ref{fig:n4} show that equivariance regularization is sensitive to team size. With $N=4$, the $N=3$ choice $\lambda_{\rm eq}=0.01$ collapses completely: $D_{\rm div}=0.25$, $F_{\rm same}=1$, and $F_{\max}=1$. A much weaker coefficient, $5\times10^{-4}$, gives the best return and improves robustness, but increases homogeneous behavior. A still weaker coefficient, $2.5\times10^{-4}$, preserves the healthiest diversity among regularized $N=4$ policies but has the worse return. This stress test supports our main claim: permutation robustness must be interpreted together with non-collapse metrics.

\begin{figure}[t]
\centering
\includegraphics[width=0.98\linewidth]{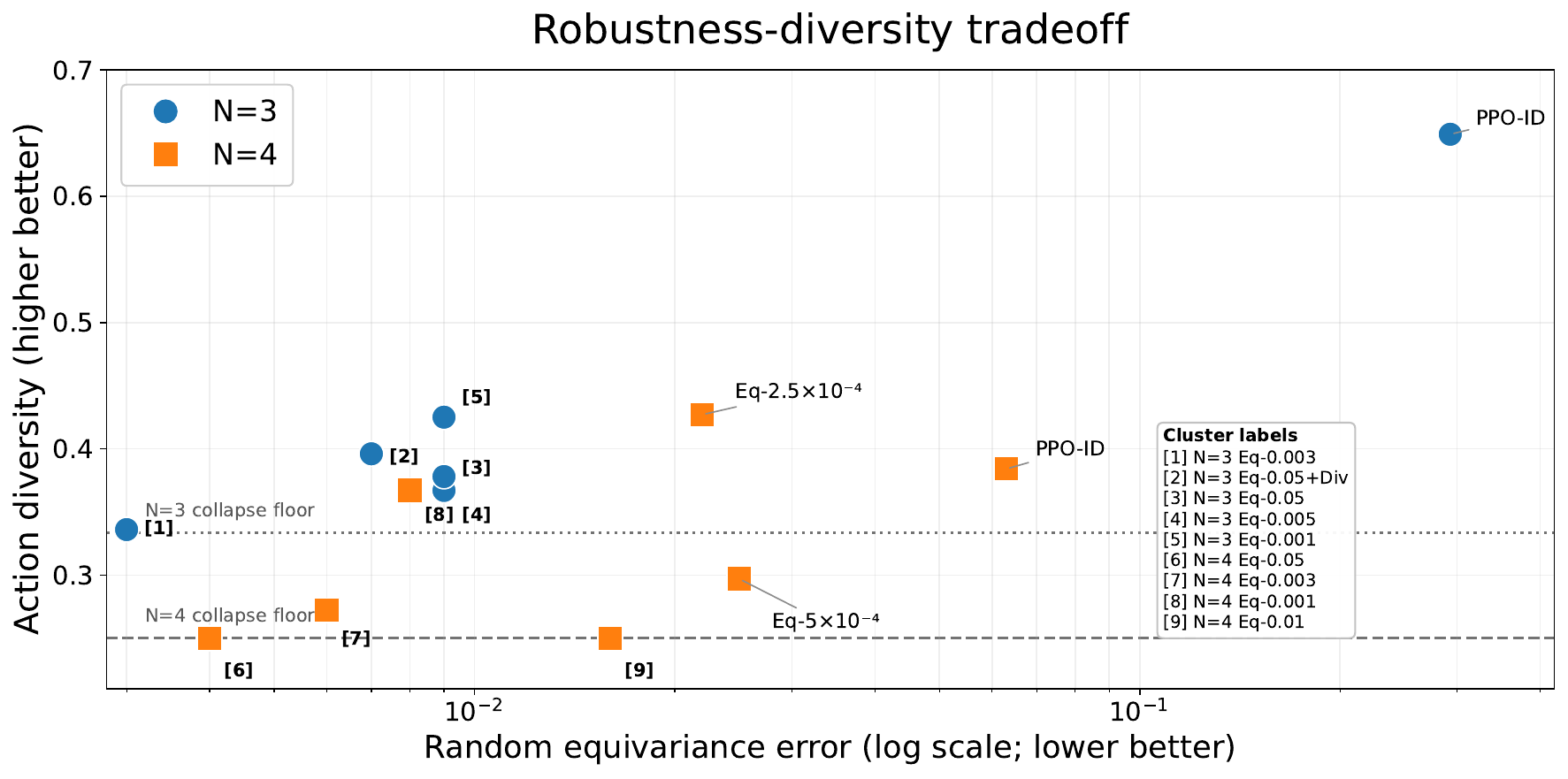}
\caption{Robustness-diversity tradeoff. The desirable region is toward the upper-left: low permutation error and high action diversity. Points near the lower-left are robust but collapse-prone. Labels are placed with offsets for readability.}
\label{fig:scatter}
\end{figure}

\section{Discussion and Limitations}
The experiments support three claims. First, transformer multi-agent policies with learned ID/position embeddings can learn strong differentiated behaviors, but those behaviors can depend on the arbitrary order in which agents are presented. Second, equivariance regularization can make policies highly permutation-consistent, but overly large regularization weights can push agents toward homogeneous action choices. Third, the useful regularization scale does not transfer naively across team sizes.

An important implication is that ``symmetry'' is not a scalar property of a learned policy. A policy may be symmetric because it uses a principled set representation, or because it has discarded the distinctions among agents. These two cases can have similar $E_{\rm eq}$ values but very different behaviors. For this reason, we recommend reporting a small metric suite rather than a single robustness score: return to measure task utility, equivariance/KL/action-disagreement metrics to measure order sensitivity, and diversity/same-action/max-frequency metrics to detect collapse. This suite also makes hyperparameter selection more defensible: we choose weak equivariance regularization not because it minimizes $E_{\rm eq}$ absolutely, but because it improves permutation consistency without fully destroying differentiated behavior.

\paragraph{Practical guidance for future evaluations:}
The results suggest a simple reporting protocol for transformer-based multi-agent robot policies. First, evaluate the policy under at least one nontrivial ordering intervention, such as random or reversed agent order. Second, report a distributional metric such as KL or logit equivariance error together with an action metric such as greedy action disagreement. Third, always include a collapse diagnostic. In our experiments, several policies obtain excellent equivariance error but have $D_{\rm div}=1/N$ and $F_{\rm same}=1$, which means all agents select the same action at every step. Such a policy should not be interpreted as having learned robust coordination.

\paragraph{Interpreting the $N=3$ and $N=4$ contrast:}
The difference between the two team sizes is especially instructive. For $N=3$, a weak coefficient $\lambda_{\rm eq}=0.01$ gives a useful regularized model: it reduces random equivariance error by more than an order of magnitude relative to \modelid, while preserving moderate action diversity. For $N=4$, the same coefficient collapses completely. Useful $N=4$ values are two orders of magnitude smaller, and even then they expose a tradeoff between return and non-collapse. This suggests that equivariance pressure effectively scales with the number of agent comparisons and the difficulty of assigning differentiated roles. In larger teams, a regularizer that is harmless for a small team may dominate the policy-gradient signal and push agents toward a shared action mode.

\paragraph{Design implications:}
Our results do not imply that equivariance is undesirable. Rather, they indicate that equivariance should be incorporated with the correct inductive bias. A model that is equivariant by construction through set pooling or relative geometric attention may avoid the need for a strong output-level penalty. Similarly, anti-collapse losses should be task-aware: in cooperative navigation, agents should spread over landmarks, but they should not be forced to choose different actions when the same motion is physically appropriate. This motivates future methods that combine permutation-equivariant architectures with assignment- or geometry-based coordination objectives.

\subsection*{Limitations}
Our experiments are limited to simulated cooperative navigation tasks. The actor is centralized and observes the joint observation. The $N=4$ experiments are a seed-0 stress test rather than a full multi-seed benchmark. Finally, pairwise action diversity is a simple anti-collapse regularizer. Future work should explore task-aware diversity, landmark assignment, relative geometric attention, and policies that preserve equivariance by construction.

\section{Conclusion}
We studied permutation sensitivity in transformer-based multi-agent robot policies and showed that robustness metrics alone can be misleading. The PPO-ID baseline produces non-collapsed behavior, but learned agent ID and positional embeddings remain order-sensitive. Equivariance regularization reduces permutation error, but can also induce same-action collapse. Therefore, evaluations of multi-agent transformer policies should report task return, permutation metrics, and action-collapse diagnostics together.

The empirical message is not that one should avoid identity embeddings or equivariance losses. Rather, these tools must be used with an explicit understanding of what is being regularized. Identity embeddings can improve specialization but create slot-specific behavior. Equivariance penalties can remove slot dependence but may also remove useful distinctions among agents. The most defensible evaluation therefore asks three questions at once: does the policy solve the task, does it behave consistently under relabeling, and does it avoid the trivial solution where all agents act the same? The broader lesson is that robust coordination requires both symmetry-aware modeling and safeguards against trivial invariant behavior.

\vspace{0.3em}
\bibliographystyle{abbrvnat}
\bibliography{references}
\clearpage

\appendix
\section{Appendix}
\label{app:appendix}

This appendix keeps the reproducibility details and mathematical explanations that support the main paper, while avoiding repeated result tables. The only tables included here are interpretation guides for reading the collapse and regularization metrics.

\subsection{Additional Experimental Details}
\label{app:details}
\paragraph{Environment:}
We use cooperative navigation from PettingZoo/MPE. Each run uses a fixed-horizon discrete-action episode and a shared team reward. The main setting has three agents and three landmarks: the stress test has four agents and four landmarks. Evaluation uses 200 episodes for each ordering mode. The appendix records these details only to support reproducibility. The main paper reports the aggregate behavior.

\paragraph{Architecture:}
The transformer policy embeds each agent observation into a 128-dimensional token. The policy uses two self-attention blocks with four heads. Agent ID embeddings and learned positional embeddings are added to the observation token. The actor head maps each final token to logits over discrete actions. The critic estimates a centralized value from the transformer representation.

\paragraph{PPO hyperparameters:}
We use rollout length 1024, four PPO epochs, minibatch size 256, learning rate $3\times10^{-4}$, discount $\gamma=0.99$, GAE parameter $\lambda=0.95$, PPO clip $\epsilon=0.2$, entropy coefficient $0.01$, value coefficient $0.5$, and max gradient norm $0.5$.

\paragraph{Permutation evaluation:}
For each observation $X$, we compute normal logits $f_\theta(X)$ and permuted logits $f_\theta(PX)$. We then unpermute the output with $P^{-1}$ before computing all comparisons. The executed action under a permuted input is also unpermuted so that the same physical agent receives the corresponding command.

\paragraph{Regularization coefficients:}
For $N=3$, the selected weak equivariance setting is $\lambda_{\rm eq}=0.01$. Stronger settings such as $\lambda_{\rm eq}=0.05$ are included to show collapse-prone behavior. For the $N=4$ stress test, useful coefficients are much smaller; the paper reports $\lambda_{\rm eq}\in\{2.5\times10^{-4},5\times10^{-4},10^{-3},3\times10^{-3},10^{-2},5\times10^{-2}\}$ to illustrate team-size sensitivity.

\subsection{Additional Mathematical Explanation}
\label{app:math}
\paragraph{Why self-attention without slot-specific embeddings is permutation equivariant:}
Let $X\in\R^{N\times d}$ be the token matrix and let $P\in\{0,1\}^{N\times N}$ be a permutation matrix. A single self-attention layer computes
\begin{equation}
Q=XW_Q,\qquad K=XW_K,\qquad V=XW_V,
\end{equation}
\begin{equation}
A=\mathrm{softmax}\!\left(\frac{QK^\top}{\sqrt{d_k}}\right),\qquad Z=AV.
\end{equation}
For the permuted input $PX$, we obtain $Q'=PQ$, $K'=PK$, and $V'=PV$. The attention scores transform as
\begin{equation}
Q'K'^\top=(PQ)(PK)^\top=P QK^\top P^\top.
\end{equation}
Because row-wise softmax preserves this simultaneous row-column permutation, $A'=PAP^\top$. Therefore,
\begin{equation}
Z'=A'V'=PAP^\top PV=PAV=PZ.
\end{equation}
Thus, self-attention is permutation equivariant when no slot-specific information is injected. Learned identity or positional embeddings can break this property because generally $PX+E\neq P(X+E)$ for a fixed embedding table $E$ tied to sequence slots.

\paragraph{Why homogeneous action collapse can fool equivariance error:}
Suppose a policy emits identical logits $z$ for every agent, independent of the observation:
\begin{equation}
    f_\theta(X)=
    \begin{bmatrix}
    z\\z\\ \vdots\\ z
    \end{bmatrix}.
\end{equation}
Then for any permutation $P$,
\begin{equation}
    P^{-1}f_\theta(PX)=f_\theta(X),
\end{equation}
so the equivariance error is zero. However, if the greedy action is also identical across agents at most timesteps, the policy has not necessarily learned differentiated coordination. This motivates reporting $D_{\rm div}$, $F_{\rm same}$, and $F_{\max}$ together with return and permutation consistency.

\paragraph{Collapse floors:}
For a team of $N$ agents, the normalized action diversity is
\begin{equation}
D_{\rm div}=\frac{1}{T}\sum_{t=1}^{T}\frac{\left|\{a_1^t,\ldots,a_N^t\}\right|}{N}.
\end{equation}
If all agents choose the same action at every timestep, then the number of unique actions is one and $D_{\rm div}=1/N$. This is the collapse floor shown in Fig.~\ref{fig:scatter}: $1/3$ for $N=3$ and $1/4$ for $N=4$.

\subsection{Additional Tables}
\label{app:additional-tables}

\begin{table}[h]
\centering
\small
\caption{Interpretation of collapse metrics.}
\label{tab:collapse-metric-guide}
\begin{tabular}{lll}
\toprule
Metric & Good direction & Failure sign \\
\midrule
$D_{\rm div}$ & high & near $1/N$ \\
$F_{\rm same}$ & low & near $1$ \\
$F_{\max}$ & low/moderate & near $1$ \\
$E_{\rm eq}$ & low & high indicates order sensitivity \\
$D_{\rm act}$ & low & high means selected actions change \\
\bottomrule
\end{tabular}
\end{table}

\end{document}